\documentclass[10pt,twocolumn,letterpaper]{article}

\usepackage[pagenumbers]{cvpr} %

\usepackage[dvipsnames]{xcolor}
\usepackage[legacy]{csvsimple}
\usepackage{multirow}
\usepackage{makecell}
\usepackage{arydshln}
\usepackage{tabularx}
\usepackage{algorithm}
\usepackage{algpseudocode}

\usepackage{graphicx}
\usepackage{duckuments}

\usepackage{subcaption}
\usepackage{caption}

\usepackage{float}

\newif\ifdraft
\drafttrue
\ifdraft
\newcommand{\gpc}[1]{{\color{blue}[\textbf{Gaurav:} #1]}}
\newcommand{\kac}[1]{{\color{purple}[\textbf{Kfir:} #1]}}
\newcommand{\opc}[1]{{\color{red}[\textbf{Or:} #1]}}
\newcommand{\dcc}[1]{{\color{orange}[\textbf{Danny:} #1]}}
\newcommand{\jacksoncomment}[1]{{\color{teal}[\textbf{Jackson:} #1]}}
\newcommand{\dosc}[1]{{\color{magenta}[\textbf{Daniil:} #1]}}

\newcommand{\jackson}[1]{{\color{teal}#1}}
\newcommand{\dos}[1]{{\color{magenta}#1}}

\else
\newcommand{\gpc}[1]{}
\newcommand{\kac}[1]{}
\newcommand{\opc}[1]{}
\newcommand{\dcc}[1]{}
\newcommand{\jacksoncomment}[1]{}
\newcommand{\dosc}[1]{}

\newcommand{\jackson}[1]{}
\newcommand{\dos}[1]{}

\fi

\newcommand{\refsec}[1]{Section~\ref{sec:#1}}

\newcommand{\lblsec}[1]{\label{sec:#1}}
\newcommand{\lbleq}[1]{\label{eq:#1}}

\newcommand{\myparagraph}[1]{\vspace{0pt}\paragraph{#1}}

\newcommand{\methodName}[0]{VisualComposer\xspace}

\definecolor{cvprblue}{rgb}{0.21,0.49,0.74}
\usepackage[pagebackref,breaklinks,colorlinks,citecolor=cvprblue]{hyperref}
\usepackage[normalem]{ulem}
\def\paperID{3337} %
\def\confName{CVPR}
\def\confYear{2025}

\title{Object-level Visual Prompts for Compositional Image Generation}
\author{
Gaurav Parmar\textsuperscript{1,2}
\qquad
Or Patashnik\textsuperscript{2,3}
\qquad
Kuan-Chieh Wang\textsuperscript{2}
\qquad
Daniil Ostashev\textsuperscript{2}
\\
Srinivasa Narasimhan\textsuperscript{1}
\qquad
Jun-Yan Zhu\textsuperscript{1}
\qquad
Daniel Cohen-Or \textsuperscript{2,3}
\qquad
Kfir Aberman\textsuperscript{2}\\
\vspace{-3.5mm}\\
\textsuperscript{1}Carnegie Mellon University\qquad
\textsuperscript{2}Snap Research\qquad
\textsuperscript{3}Tel Aviv University\\
\url{https://snap-research.github.io/visual-composer/}
\vspace{4mm}
}

\begin{document}
\twocolumn[{%
\renewcommand\twocolumn[1][]{#1}%
\vspace{-1em}
\maketitle

\vspace{-1em}
\begin{center}
    \centering 
    \vspace{-0.3in}
    \includegraphics[width=\linewidth]{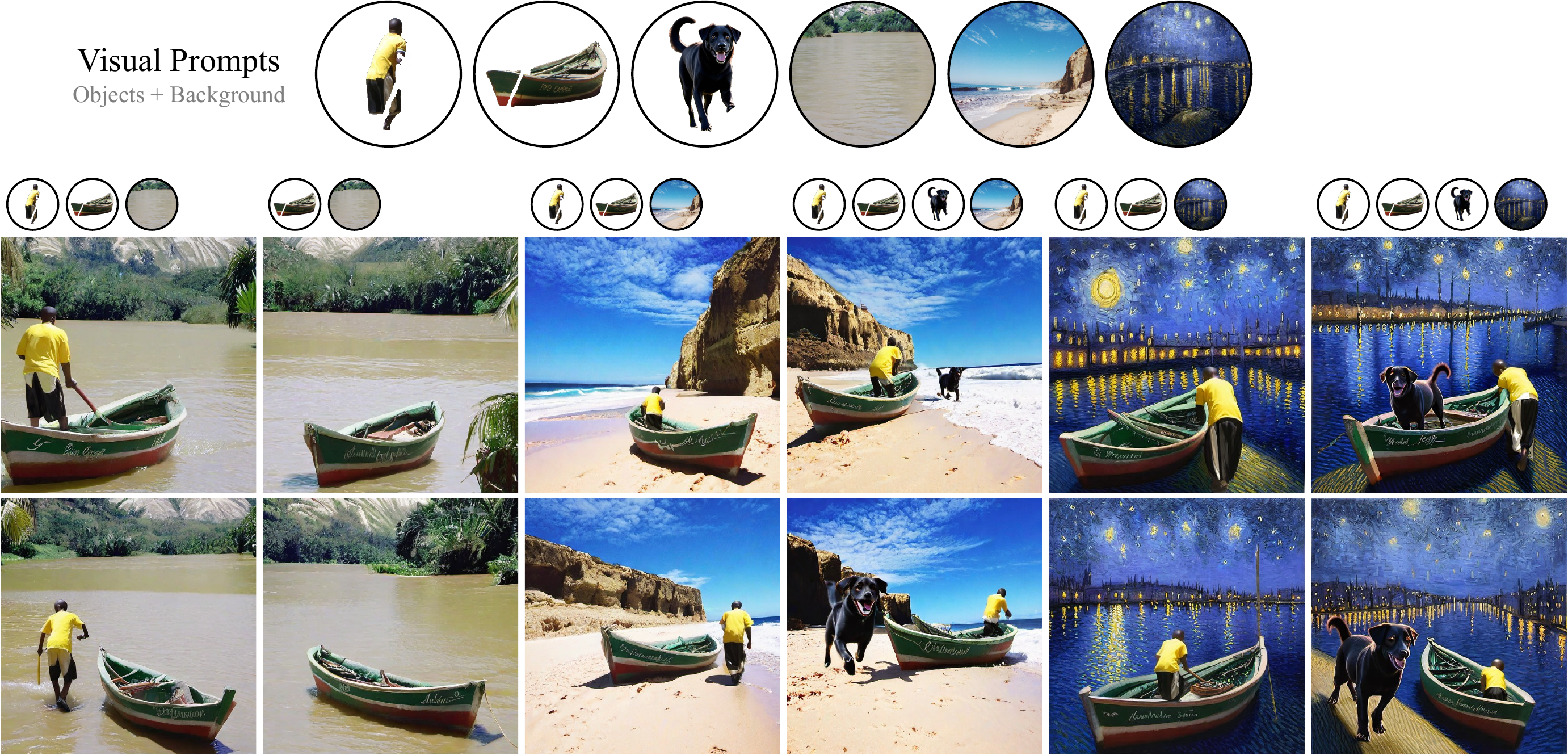}
    \vspace{-6mm}
    \captionof{figure}{
We introduce a method for composing object-level visual prompts (shown above each column), where prompts consist of both foreground and background elements that jointly guide the generation in text-to-image models. Similar to text prompts, these visual prompts enable creating semantically coherent compositions across a variety of styles and scenes without the need for a predefined layout.
    }
    \label{fig:teaser}
\end{center}%
}]

\begin{abstract}

We introduce a method for composing object-level visual prompts within a text-to-image diffusion model. Our approach addresses the task of generating semantically coherent compositions across diverse scenes and styles, similar to the versatility and expressiveness offered by text prompts.
A key challenge in this task is to preserve the identity of the objects depicted in the input visual prompts, while also generating diverse compositions across different images.
To address this challenge, we introduce a new KV-mixed cross-attention mechanism, in which keys and values are learned from distinct visual representations. The keys are derived from an encoder with a small bottleneck for layout control, whereas the values come from a larger bottleneck encoder that captures fine-grained appearance details. 
By mixing keys and values from these complementary sources, our model preserves the identity of the visual prompts while supporting flexible variations in object arrangement, pose, and composition. 
During inference, we further propose object-level compositional guidance to improve the method's identity preservation and layout correctness. 
Results show that our technique produces diverse scene compositions that preserve the unique characteristics of each visual prompt, expanding the creative potential of text-to-image generation.

\end{abstract}
  
\vspace{-5mm}
\section{Introduction} \label{sec:intro}

Text-to-image models~\cite{rombach2022high, saharia2022photorealistic, ho2020denoising, kang2023gigagan} have made remarkable progress, enabling photorealistic image synthesis with a wide variety of object compositions and arrangements. These models can create complex scenes with multiple interacting elements that generally align with user-provided textual prompts. However, integrating visual prompts, which are images that guide the generation process, is not a native capability of common model architectures, which lack an inherent mechanism for using them to generate semantically coherent compositions.
As a result, personalization and customization methods have emerged to address this limitation~\cite{ruiz2022dreambooth, gal2022textual, kumari2022customdiffusion}. Initial methods required per-subject optimization, adding a significant computational overhead per subject. Recently, feed-forward methods were introduced to accelerate the process.
One widely used method is image prompt adapters (IP-Adapters)~\cite{ye2023ipadapter}. 
These adapters encode the entire input image and incorporate it into the model through decoupled cross-attention layers, allowing it to process textual and visual cues jointly.
 
\begin{figure}[t!]
    \centering 
    \includegraphics[width=\linewidth]{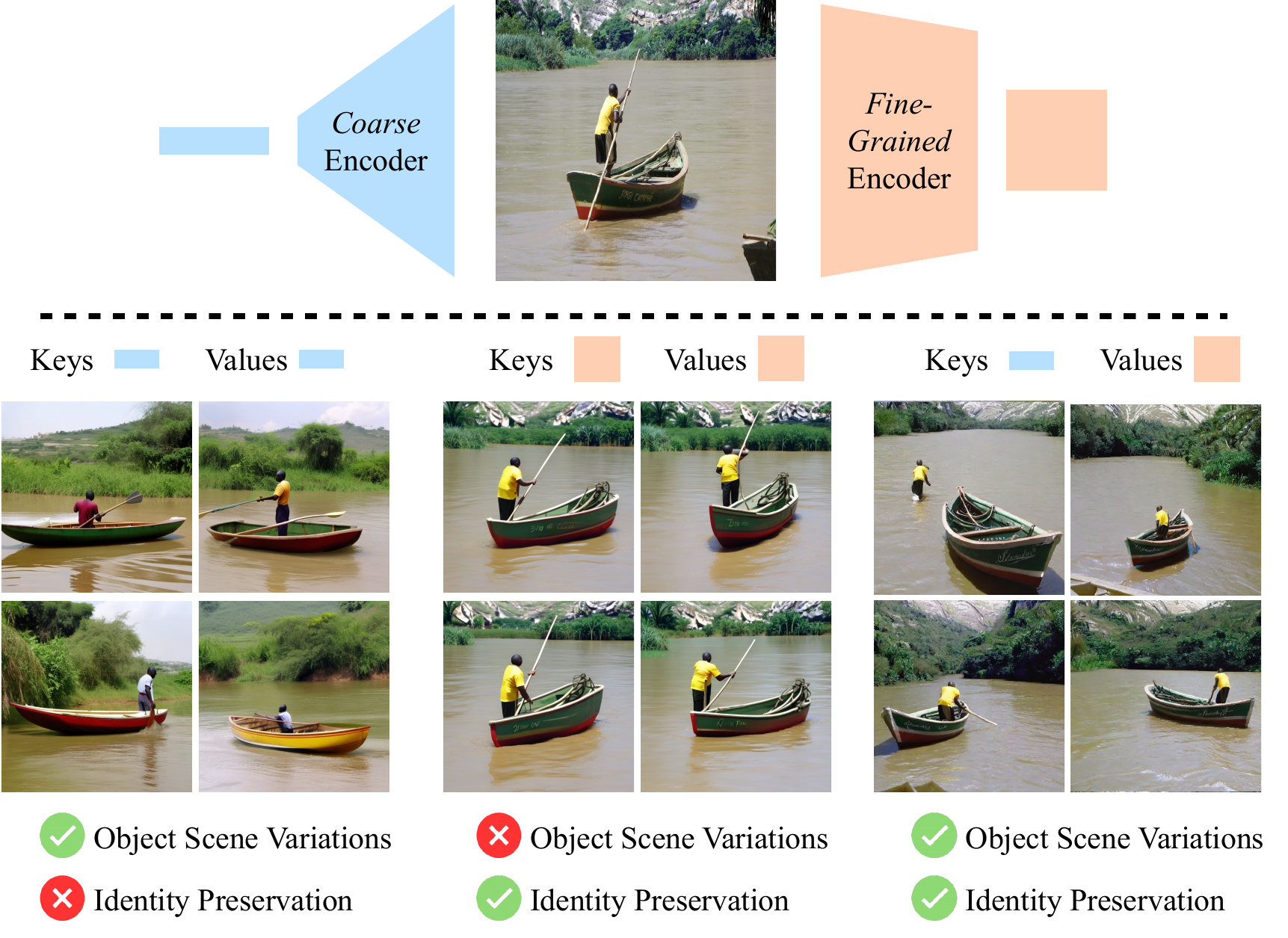}
    \begin{subfigure}{0.33\linewidth}
        \centering
        \caption{Coarse KV}
        \label{fig:coarse-kv}
    \end{subfigure}%
    \begin{subfigure}{0.33\linewidth}
        \centering
        \caption{Fine KV}
        \label{fig:fine-kv}
    \end{subfigure}%
    \begin{subfigure}{0.33\linewidth}
        \centering
        \caption{Mixed KV (Ours)}
        \label{fig:mixed-kv}
    \end{subfigure}
    \vspace{-2mm}
    \caption{\textbf{KV-Mixing.} Image Prompt Adapters capture visual information from images to guide the generation process. The feature extractor's bottleneck size (top row) determines the level of detail in the extracted Key-Value (KV) features. 
    Using only coarse KVs (left) sacrifices identity preservation, while using only fine-grained KVs (middle) limits scene variation. In contrast, combining mixed-granularity KVs (right) achieves diverse scene representation without compromising identity preservation.}
    \label{fig:kv_mix_motivation}
    \vspace{-5mm}
\end{figure}

While IP-Adapters offer additional control, they present two main drawbacks. First, they treat the input image as a single, unified prompt, limiting the model’s ability to differentiate and control individual objects within the scene. Second, these adapters encounter an inherent \textit{identity-diversity tradeoff} when balancing the identity preservation of the objects depicted in the visual prompts with diversity in the generated compositions. As shown in Figure \ref{fig:kv_mix_motivation}, an adapter with a small bottleneck (left) struggles to preserve object identity, resulting in a loss of detail. Conversely, a large bottleneck (middle) improves the identity preservation of the prompt image but overfits to its structure, leading to limited variation in layouts and poses. These limitations highlight the need for a method that generates coherent, flexible compositions while preserving the distinct characteristics of individual visual elements.

In this paper, we present a novel technique for generating coherent compositions by incorporating object-level visual prompts into text-to-image diffusion models (see the gallery of compositions in Figures \ref{fig:teaser} and~\ref{fig:results}).
Our technique addresses the identity preservation-diversity trade-off and enables versatile image compositions. 
Our approach begins by examining the distinct roles of keys and values extracted from the image prompt, where the keys control the layout of the generated scene and the values encode the fine-grained appearance details~\cite{hertz2022prompt, parmar2023zero, tewel2023keylocked, Cao_2023_ICCV}. Building on this insight, we propose a KV-mixed cross-attention module that leverages two encoders, one with a small bottleneck (global) image encoder for the keys and one with a larger bottleneck (local) image encoder for the values. This cross-attention module is referred to as ``KV-mixed'', as it mixes keys and values learned from the two distinct visual representations. Furthermore, we propose Compositional Guidance, an object-level guidance method to improve identity preservation and layout coherence during inference time. 

Our method provides a powerful framework for generating diverse compositions from a defined set of visual elements, balancing identity preservation with adaptability across various layouts and poses. Our results demonstrate that this approach yields coherent, detail-rich images while maintaining flexibility in object arrangement and scene composition. Our method outperforms image prompting, optimization methods, and multi-modal generation methods on the compositional image generation benchmark.

\begin{figure*}[t!]
    \centering 
    \includegraphics[width=0.95\linewidth]{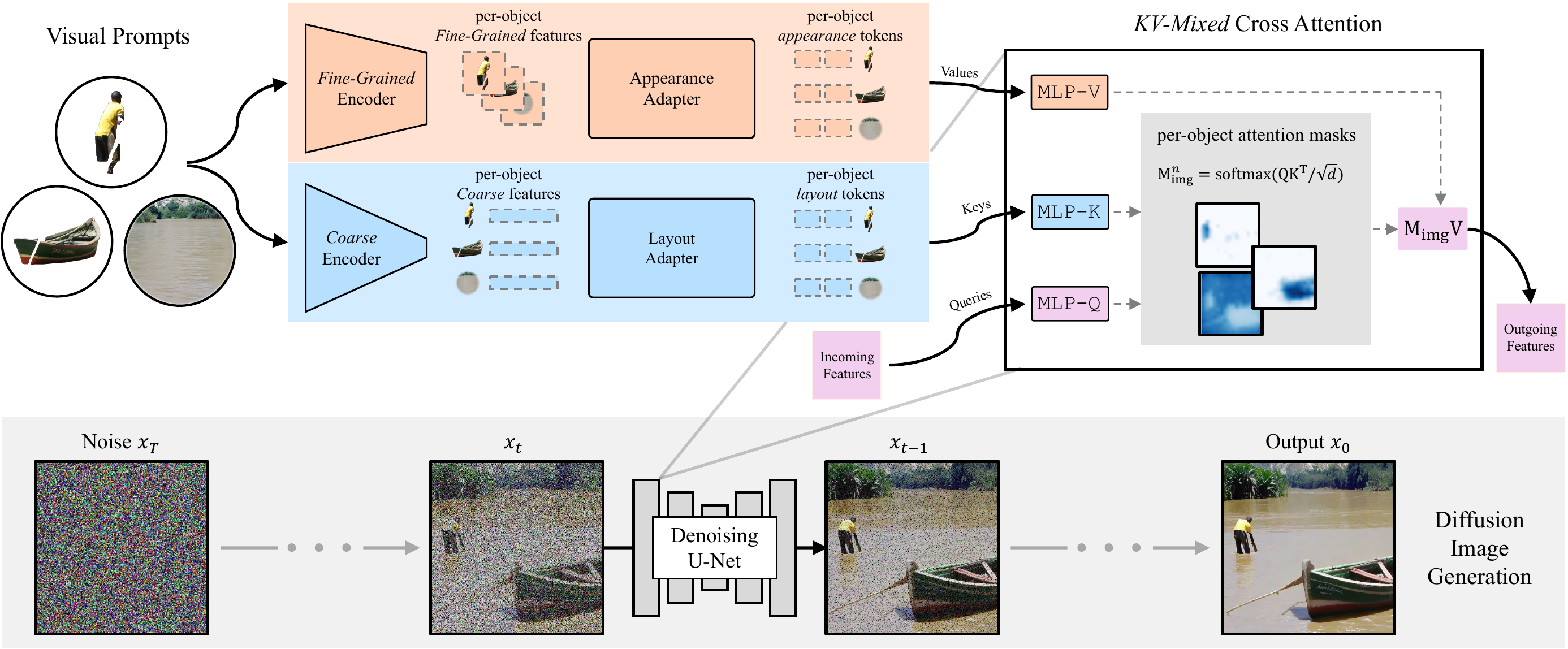}
    \caption{ 
    \textbf{\methodName architecture.} 
    Our method begins by encoding all input visual prompts through two separate branches: an appearance branch (top row, shown in orange) that uses a Fine-Grained encoder followed by an Appearance adapter to encode \emph{per-prompt} appearance tokens, and a layout branch (bottom row, shown in blue) that uses a Coarse encoder followed by a Layout adapter to encode \emph{per-prompt} layout tokens. Once the appearance and layout tokens are extracted from the input visual prompts, they are injected into the U-Net through Object-Centric KV-Mixed Cross Attention layers. The layout tokens are input as keys and determine the spatial influence of each individual visual prompt in the final image, as visualized by the per-object attention masks. The appearance tokens are input as values \emph{after} attention mask is computed and hence \emph{only} influence the appearance and the identity. 
    } 
    \label{fig:method_pipeline}
\end{figure*}

\section{Related Works} \label{sec:related_works}

\myparagraph{Single-concept personalization.}
Recent large-scale image generative models typically rely on text prompts for conditioning~\cite{rombach2022high, podell2024sdxl, ramesh2022hierarchical, saharia2022photorealistic, kang2023gigagan}. While text provides an intuitive interface for image synthesis, its expressiveness is limited when describing specific visual elements. To address this limitation, numerous works have developed means to embed images into the model~\cite{gal2022textual, ruiz2022dreambooth, kumari2022customdiffusion, avrahami2023bas}, thereby enabling image-based conditioning for synthesis. Initial approaches required per-subject optimization, which restricted their applicability due to high computational costs. More recent works have focused on training encoders or adapters to condition the generation on input images in a feed-forward manner~\cite{gal2023encoder, arar2023domain, ruiz2023hyperdreambooth, ye2023ipadapter, zeng2024jedi, Wei_2023_ICCV, shi2023instantbooth, chen2023subjectdriven, jia2023tamingencoderzerofinetuning}. 

\vspace{-12pt}
\myparagraph{Multiple-subject scene generation.}
Generating complex scenes with multiple interacting objects presents a substantial challenge~\cite{chefer2023attendandexcite, Phung_2024, seperate_enhance, dahary2024yourself, li2023gligen, wang2024moa, gu2024mix}. As a result, most encoder-based personalization methods focus on a single object~\cite{zeng2024jedi, arar2023domain, Wei_2023_ICCV, shi2023instantbooth, chen2023subjectdriven, jia2023tamingencoderzerofinetuning, li2023blipdiffusion}. 
To address the difficulty of generating scenes with multiple subjects, researchers have developed several dedicated methods. For instance, certain methods~\cite{kumari2022customdiffusion, po2023orthogonal} merge separately learned concepts within a single image. Break-a-Scene~\cite{avrahami2023bas} takes a different approach by assuming the existence of an image containing all objects, from which it learns separate representations for each object. However, these methods require a lengthy optimization process for each object or scene. Alternatively, other methods~\cite{xiao2023fastcomposer, wang2024moa, kim2024instantfamily} enable feed-forward multi-subject generation but are limited to human faces and cannot handle general objects.

A significant challenge in generating novel images of objects is balancing the preservation of the object's appearance with the diversity of the generated images~\cite{alaluf2023neural, tewel2023keylocked}. As shown in Figure~\ref{fig:kv_mix_motivation}, methods that excel at faithfully preserving object appearance often struggle to generate diverse layouts. %
This issue is especially noticeable in scenes with multiple objects, resulting in repetitive image compositions. Conversely, approaches that prioritize diversity often struggle to maintain the precise appearance of the original objects. We design our method to generate diverse images, especially in terms of their composition, while simultaneously preserving the object appearance depicted in the image prompt.

\vspace{-2mm}
\myparagraph{Layout-conditioned scene generation.} Another approach to addressing the complexity of multi-object generation relies on input layouts, either using conditional diffusion models~\cite{Chen_2024, kim2024instantfamily, zhang2023adding, li2023gligen,parmar2024one,avrahami2023spatext,wang2024instancediffusion} or leveraging training-free inference methods~\cite{chen2024training,kim2023dense,he2023localized,dahary2024yourself,phung2024grounded}. While this approach facilitates multi-object generation, it presents two main challenges. First, it requires users to provide a layout compatible with the text prompt, constraining the diversity of generated images to the provided layout. Second, this approach often limits the level of interaction between generated objects, impacting the cohesiveness of the final scene.

\vspace{-2mm}
\section{Method} \label{sec:method}
\vspace{-2mm}
Given a set of $N$ input visual prompts $\{\mathcal{P}_v^n\}_{n=1}^{N}$ describing the $N-1$ individual objects and the background of an image, our goal is to generate diverse output images composed of these inputs.
We first discuss text-to-image diffusion models and image encoder preliminaries in \refsec{prelim}. 
Following this, \refsec{arch} explores the trade-off between maintaining the identity of input elements and introducing variation in the generated images, which motivates our architecture design. \refsec{training} details our training method and datasets, and lastly, \refsec{inference} describes our new compositional guidance for inference.
We refer to our method as \methodName.

\subsection{Preliminaries} \label{sec:prelim}
\myparagraph{Text-to-Image Diffusion.}
Diffusion models~\cite{sohl2015deep, ho2020denoising, songscore} are a family of generative models that use iterative denoising processes. Recent diffusion models are typically conditioned on text prompts~\cite{rombach2022high, podell2024sdxl} through cross-attention layers~\cite{bahdanau2015neural}. 
Specifically, a text embedding vector $c$ is derived from a text prompt $\mathcal{P}_t$ using a frozen CLIP~\cite{radford2021learning} text encoder $c = E_\text{text}(\mathcal{P}_t)$. 
This text embedding interacts with the generated image deep spatial features $\phi(x_t)$ as follows.  
The image features $\phi(x_t)$ are projected to queries $Q = f_Q(\phi(x_t))$, while the text embedding is projected to keys $K = f_K (c)$ and values $V = f_V (c)$, where $f_Q, f_K, \text{and } f_V$ are learned linear layers. The output of the cross-attention layer is computed as $\mathcal{M}V$, where $\mathcal{M}$ are the attention maps defined as $\mathcal{M} = \text{Softmax}\left( QK^T/\sqrt{d} \right).$

Previous works have shown that each component of attention has its own role~\cite{hertz2022prompt, parmar2023zero, tewel2023keylocked, alaluf2024cia}. The keys, which form the attention map, tend to control the layout, and the values determine the appearance. We use this observation to control the identity preservation-diversity tradeoff.
 
\vspace{-14pt}
\myparagraph{Prompting with images.}
While natural language allows us to control generation with simple words, it often fails to provide precise descriptions of objects. Recent methods~\cite{ye2023ipadapter,gal2023encoder,ruiz2023hyperdreambooth} extend text-to-image diffusion models to also condition on image prompts.
For example, in IP-Adapter~\cite{ye2023ipadapter}, the image prompt $\mathcal{P}_\text{img}$
is first encoded with a pretrained image encoder to obtain image embeddings $E_\text{img}(\mathcal{P}_\text{img})$, and then transformed through a learned adapter network $A$  to form the image tokens $c_{\text{img}} = A(E_\text{img}(\mathcal{P}_\text{img}))$.
Next, the image tokens are projected to the corresponding image prompt keys $K_\text{img} = f_K^\text{img} (c_{\text{img}})$ and values $V_\text{img} = f_V^\text{img} (c_{\text{img}})$, with new image prompt linear layers $f_K^\text{img}$ and $f_V^\text{img}$. 
Analogous to text prompts, the image prompt attention map is defined as  $\mathcal{M}_\text{img} = \text{Softmax}\left( QK_{\text{img}}^T/\sqrt{d} \right).$

The output of the decoupled cross-attention~\cite{ye2023ipadapter} is computed as a sum of text-prompt cross-attention and image-prompt cross-attention $\mathcal{M}V + \mathcal{M}_\text{img}V_\text{img}$.
Different methods~\cite{gal2023encoder, arar2023domain, ruiz2023hyperdreambooth, ye2023ipadapter, zeng2024jedi, shi2023instantbooth} differ in encoder designs,  image embedding dimensions, and adapter architectures.

\subsection{The \methodName Architecture}
\lblsec{arch}

\myparagraph{Exploring the identity preservation-diversity tradeoff.}
As discussed in Section~\ref{sec:prelim}, prompting with images typically begins by extracting image features $E_\text{img}(\mathcal{P}_\text{img})$, where $E_\text{img}$ is a pre-trained frozen image encoder.
We find that the choice of feature extractor is crucial, as it impacts the trade-off between identity preservation and output diversity.
Encoders with a narrow information bottleneck (\ie, heavy information compression) may not capture sufficient details about the object's identity, but they tend to generate more diverse results as they are less likely to overfit the original pose or spatial arrangement. 
In contrast, encoders with a wide information bottleneck (\ie, retaining highly detailed information) better capture the identity features but tend to overfit to the original pose and layout, thereby sacrificing the model's ability to generalize to new poses.

This observation is shown in Figure~\ref{fig:kv_mix_motivation}. 
There, we encode the image on the top row using two types of encoders, one with a narrow bottleneck and the other with a wide bottleneck. 
The image embeddings from each encoder are processed through their corresponding pretrained IP-Adapter~\cite{ye2023ipadapter} and injected into the diffusion model via decoupled cross-attention layers, resulting in the generation of four images.
In Figure~\ref{fig:coarse-kv}, the narrow bottleneck encoder features result in diverse layouts but poor identity preservation. Conversely, in Figure~\ref{fig:fine-kv}, the wide bottleneck encoder features preserve identity but suffer from layout overfitting.

\vspace{-3mm}
\myparagraph{KV-Mixed Cross-Attention.} 
Our method overcomes the identity preservation-diversity tradeoff by leveraging the unique roles of keys and values in the cross-attention mechanism. We introduce KV-Mixed Cross-Attention Layers, where we employ a coarse (narrow bottleneck) encoder $E_\text{img}^{C}$ for the keys to promote diversity in poses and layouts, while a fine-grained (wide bottleneck) encoder $E_\text{img}^{F}$ is used for the values to preserve detailed identity features accurately.
As shown in Figure~\ref{fig:mixed-kv}, by \textit{mixing} the features obtained from the two encoders, we are able to achieve high identity preservation of the input image while also generating diverse layouts.

\vspace{-3mm}
\myparagraph{Architecture.}
Our method's architecture is illustrated in Figure~\ref{fig:method_pipeline}. Given the $N$ visual prompts $\{\mathcal{P}_v^n\}_{n=1}^{N}$ shown on the left, we generate images that preserves the identity of each prompt while allowing for flexible layouts and poses.

Our method builds upon a pre-trained text-to-image diffusion model~\cite{rombach2022high, podell2024sdxl}, which remains frozen during training. Each input visual prompt, $\mathcal{P}_v^n$, is processed through a two-stream architecture. 
The first stream (top of Figure~\ref{fig:method_pipeline}) uses a fine-grained encoder $E_\text{img}^F$ followed by a transformer and an appearance adapter $A_\text{app}$ to extract appearance tokens $\{A_\text{app} (E_\text{img}^F (\mathcal{P}_v^n)) \}_{n=1}^{N}$. 
The second stream (bottom) utilizes a coarse encoder $E_\text{img}^C$ followed by a transformer and a layout adapter $A_\text{layout}$ to obtain layout tokens $\{A_\text{layout} (E_\text{img}^C (\mathcal{P}_v^n))\}_{n=1}^{N}$.

The fine-grained encoder is implemented using a CLIP image encoder, extracting grid features from its penultimate layer.
The coarse encoder uses the CLIP global image embedding. 
The appearance adapter is implemented as a Perceiver Transformer ~\cite{jaegle2021perceiver}, and the layout adapter is implemented as a linear layer with layer normalization \cite{ba2016layer}.
Extracted layout and appearance tokens from each visual prompt are concatenated and fed to our KV-Mixed cross-attention layers, serving as keys and values, respectively. These decoupled KV-Mixed cross-attention layers are added to each cross-attention layer.

\subsection{Training} \label{sec:training}
\myparagraph{Dataset.}
Our training dataset combines real images~\cite{kakaobrain2022coyo-700m} and synthetically generated multi-object images~\cite{flux}. Each training sample consists of an input image $x$, a text prompt, and a set of $N-1$ binary object masks $\{m_n\}_{n=1}^{N-1}$. The sample also includes a background image $x_\text{bg}$, obtained by inpainting all masked objects in $x$. We define the object visual prompts $\{\mathcal{P}_v^n\}_{n=1}^{N-1}$ by applying each mask $m_n$ to the original image $x$. The $N$-th visual prompt $\mathcal{P}_v^N$ corresponds to the background image $x_\text{bg}$. For additional implementation details, please refer to the Appendix.

\vspace{-15pt}
\myparagraph{Objective.}
Given the visual prompts $\{\mathcal{P}_v^n\}_{n=1}^{N}$ and the text prompt, we train our method to reconstruct the image $x$.
During training, we optimize only the following components: the appearance adapter $A_\text{app}$, the layout adapter $A_\text{layout}$, and the linear layers $f_K^{\text{img}}, f_V^{\text{img}}, f_Q$ of the KV-Mixed cross-attention layers, while keeping the base text-to-image diffusion model and image encoders $E_\text{img}^F, E_\text{img}^C$ frozen.
Our training objective combines two losses. The first is the standard diffusion reconstruction loss. The second is a bounded cross-attention loss $\mathcal{L}_\text{xa}$ that encourages alignment between the KV-mixed cross-attention maps $\mathcal{M}_\text{img}^n$ and their corresponding binary object masks $m_n$~\cite{xiao2023fastcomposer, dahary2024yourself}. Specifically, $\mathcal{L}_\text{xa}$ penalizes attention given to visual prompt $\mathcal{P}_v^n$ in regions of the target image $x$ where the object does not appear. 
For each visual prompt $\mathcal{P}_v^n$ that corresponds to an object, we define:
\begin{equation}\lbleq{xa_loss}
    \begin{aligned}
        \mathcal{L}_\text{xa}^{n} = 1 - \frac{ \mathcal{M}_\text{img}^n \odot m_n  }{\mathcal{M}_\text{img}^n \odot m_n + \alpha \mathcal{M}_\text{img}^n \odot (1-m_n)},
    \end{aligned}
\end{equation} 
where $\odot$ denotes the Hadamard product, and $\alpha$ is a hyperparameter that controls the significance of the background regions. $\mathcal{L}_\text{xa}$ is then defined as $\sum_{n=1}^{N-1} {\mathcal{L}_\text{xa}^{n}}$.

\begin{figure*}[ht!]
    \centering 
    \includegraphics[width=0.92\linewidth]{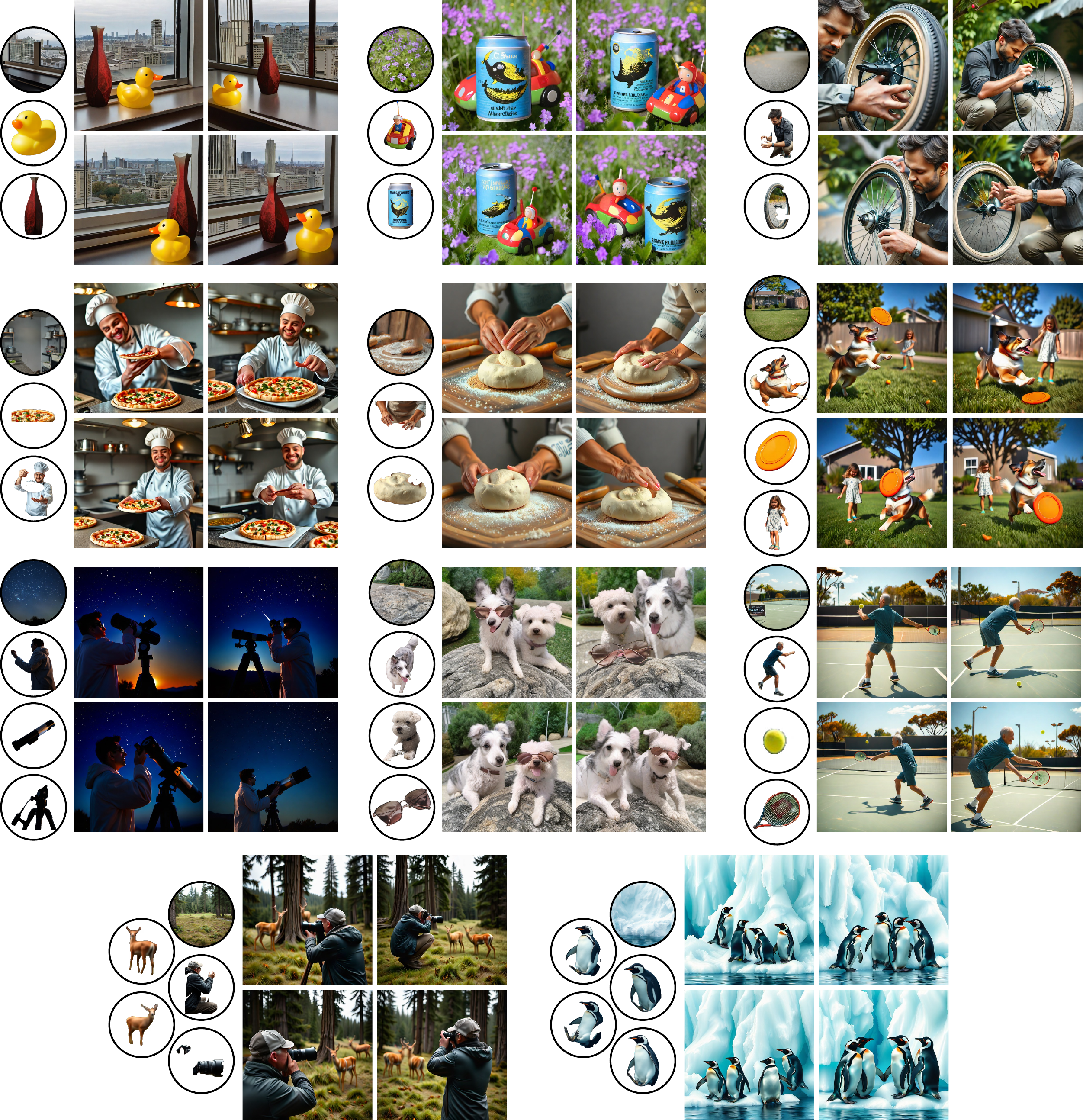}
    \vspace{-2mm}
    \caption{ \textbf{Gallery.} Compositional images generated by \methodName. Four outputs (right) for each set of input visual prompts (left).
    } 
    \label{fig:results}
    \vspace{-5mm}
\end{figure*}

\subsection{Inference} \label{sec:inference}
\vspace{-2mm}
\noindent
Previous works~\cite{chefer2023attendandexcite, seperate_enhance, phung2024grounded, dahary2024yourself} have shown that text-to-image models often struggle to faithfully follow input text prompts. These works use guidance-based inference-time techniques~\cite{dhariwal2021diffusion, ho2021classifier} to improve text-image alignment. 
To enhance the model's adherence to input \textit{visual} prompts, we introduce Compositional Guidance during inference.
This technique relies on individual object segments in the generated image. To this end, our generation process is done in two stages. First, we generate an image without any intervention in the denoising process and find a segment for each visual prompt. 
These segments are used to generate the final output image as described below. 

\vspace{-4mm}
\myparagraph{Assigning Segments.}
We begin by applying an open-set segmentation~\cite{ravi2024sam2, yuan2021florence} on the image generated in the first stage, and denote by $\{\mathcal{S}_j\}$ the set of detected segments.
Then, to match each visual prompt with a segment, we use an optimal assignment algorithm, where
we compute the DINOv2 ~\cite{oquab2023dinov2} similarity between each input visual prompt $\mathcal{P}_v^n$ and the detected segment $S_j$:
\vspace{-2mm}
\begin{equation} \label{eq:dino-sim}
    \text{Sim}(n, j) = \text{DINO}(\mathcal{P}_v^n, S_j).
\end{equation}
\vspace{-1mm}
Using DINOv2 similarity as the cost function (computed as $1 - \text{Sim}(n, j)$), we use the Hungarian matching algorithm \cite{kuhn1955hungarian} to find the best one-to-one assignment $\sigma(n)$ between the input visual prompts and the detected segments. 

\vspace{-4mm}
\myparagraph{Compositional Guidance.}
To reinforce the correspondence between input visual prompts and their generated counterparts, we adjust the attention maps during inference. For each input visual prompt $\mathcal{P}_v^n$, we modify its associated attention map $\mathcal{M}_\text{img}^{n}$ by zeroing out values outside the region of the matched segment $S_{\sigma(n)}$. Specifically, we do so by setting these values as $-\infty$ in the result of $QK_\text{img}^T$ before applying the Softmax that produces $\mathcal{M}_\text{img}^{n}$.
We further define a loss function to maximize the DINO similarity between each input visual prompt and its matched segment: 
\vspace{-3mm}
    \begin{equation}
        \mathcal{L}_\text{id} = \sum_{n} ( 1 - \text{Sim}(n, \sigma(n) ) ),
    \end{equation}
\vspace{-1mm}
where Sim is defined as in Equation~\ref{eq:dino-sim}. Note, that here the similarity is computed between the input visual prompt $\mathcal{P}_v^n$ and the segmentation mask of $S_{\sigma(n)}$ applied on the $x_0$ prediction of the current noisy image $z_t$.
We backpropagate this loss through the model to update the appearance tokens of the fine-grained encoder $A_\text{app} (E_\text{img}^F (\mathcal{P}_v^n)) $. Updating only the appearance tokens ensures that only the identity features are refined without affecting the overall scene layout.

\begin{figure*}[ht!]
    \centering 
    \includegraphics[width=0.9\linewidth]{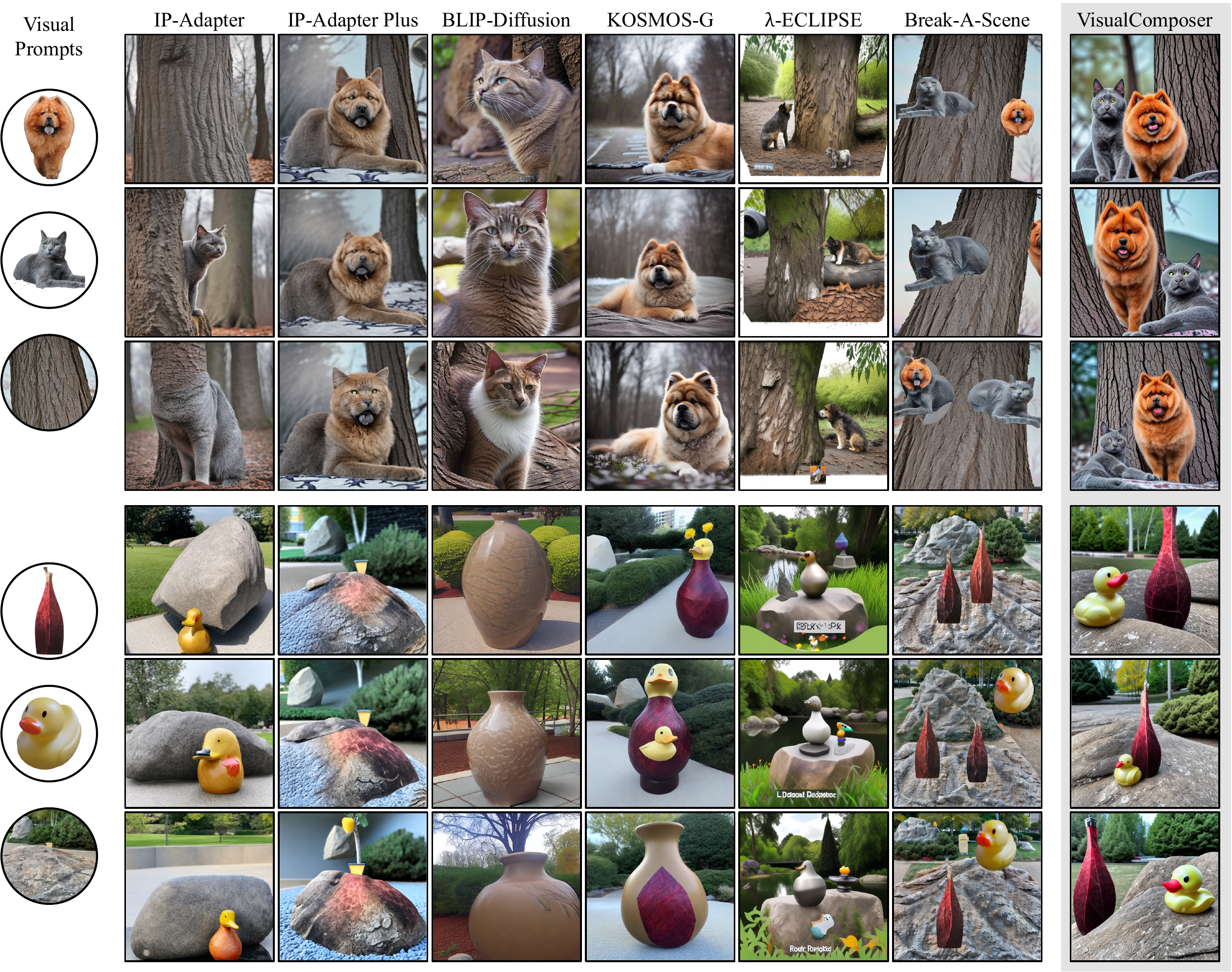}
    \vspace{-2mm}
    \caption{ \textbf{Comparisons to prior methods.} We show a set of input visual prompts on the left. For each set, we show results generated by different methods. Our method achieves the best balance between identity preservation of the input prompts and image diversity. Our method is the only one that successfully generates the two objects in realistic layouts without fusing them or outputting duplicates. } 
    \label{fig:comparisons}
    \vspace{-2mm}
\end{figure*}

\section{Experiments} \label{sec:experiments}
\begin{figure}[ht!]
    \centering 
    \includegraphics[width=\linewidth]{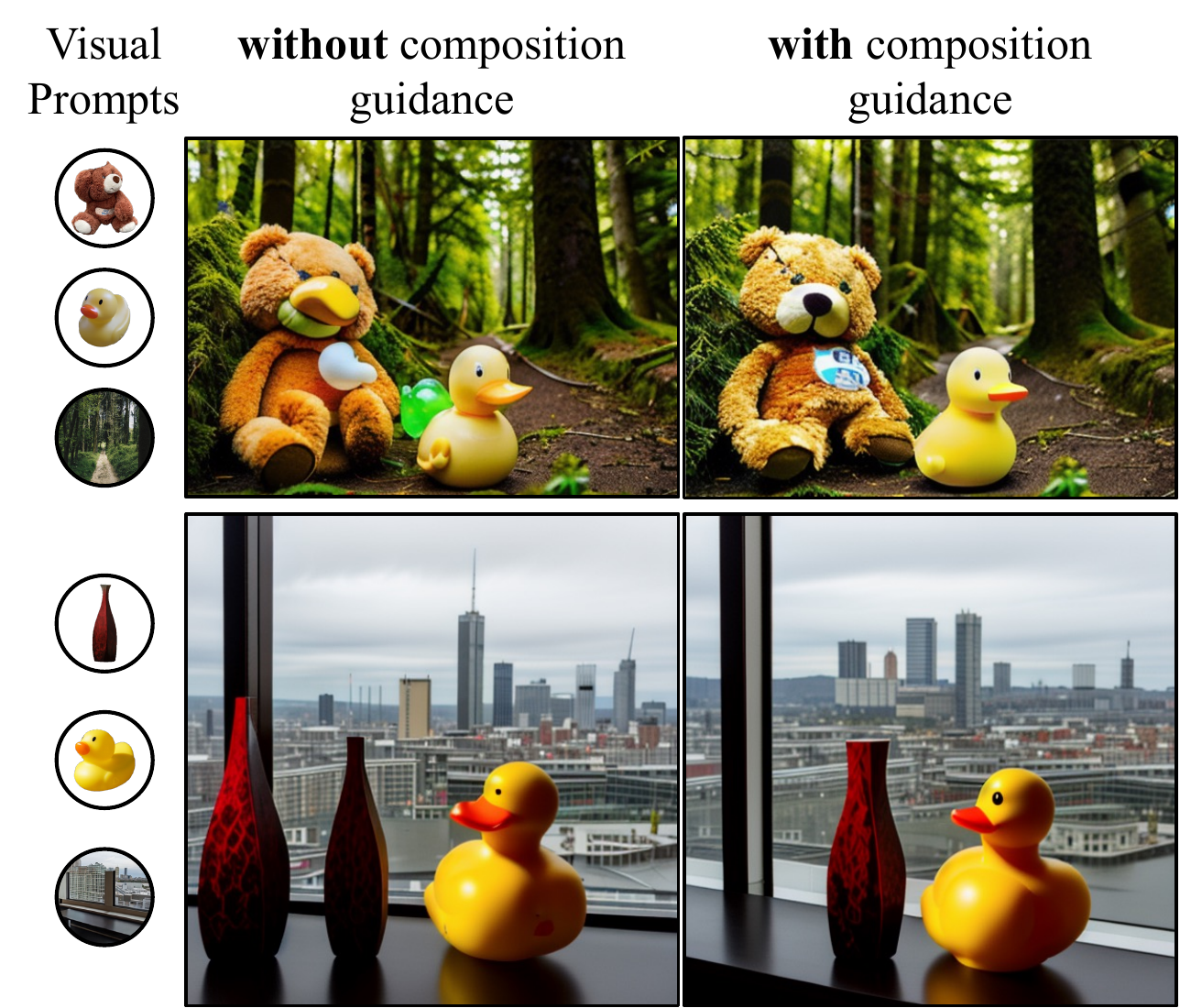}
    \vspace{-5mm}
    \caption{\textbf{Ablating Compositional Guidance.} 
    Our inference-time compositional guidance improves identity preservation, reduces leakage between objects, and removes duplicates. Without guidance, the duck's features leak into the bear (top row) and two vases get generated (bottom row).}
    \label{fig:ablations}
\end{figure}

\begin{figure}[ht!]
    \centering 
    \includegraphics[width=\linewidth]{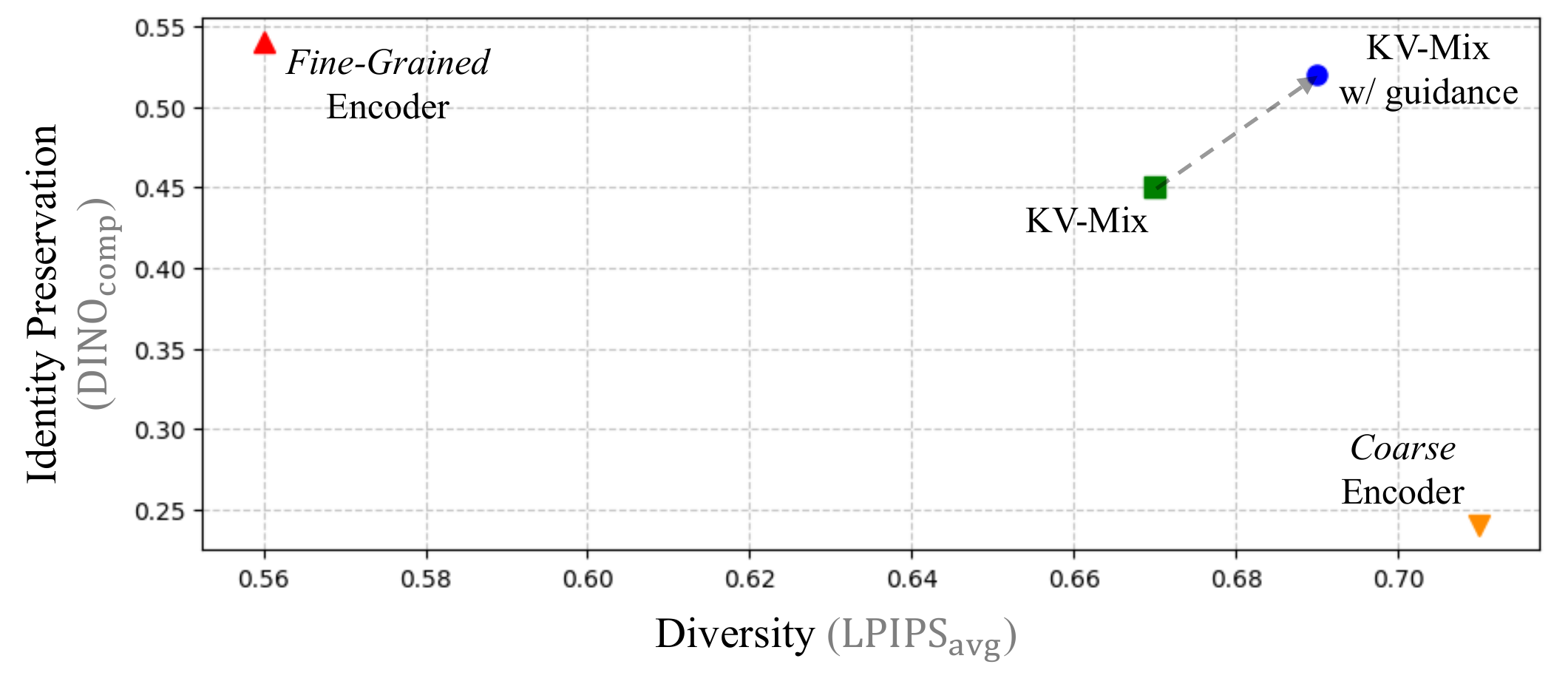}
    \vspace{-6mm}
    \caption{\textbf{Ablations.} We ablate the importance of KV-Mixture and Compositional guidance. If Fine-Grained encoder is used for both keys and values, the method overfits and does not generate adequate variations. Conversely, using Coarse encoder results in poor identity preservation. Using KV-Mixture achieves a better trade-off, and is further improved with compositional guidance. }
    \label{fig:plt_ablation}
    \vspace{-2mm}
\end{figure}

In this section, we demonstrate the effectiveness of our method through a series of experiments. Section~\ref{sec:eval_protocol} begins by discussing the evaluation protocol used. 
Section~\ref{sec:cmp_baselines} shows how our method compares with previous approaches, and Section~\ref{sec:cmp_ablations} demonstrates the importance of each individual component of our method. 
We train our method using Stable Diffusion 1.5~\cite{rombach2022high} and Stable Diffusion XL~\cite{podellsdxl} as the base text-to-image diffusion models. 
For a fair comparison to the baseline methods, Figures~\ref{fig:comparisons},~\ref{fig:ablations},~\ref{fig:plt_ablation}, and Table~\ref{tab:cmp_baselines_combined} use the Stable Diffusion 1.5 model. 
The results in Figures~\ref{fig:teaser} and~\ref{fig:results} use Stable Diffusion XL. 
A classifier free guidance value of 7.5 and the DDIM scheduler \cite{song2022denoisingdiffusionimplicitmodels} with 25 denoising inference steps are used in all comparisons. 
Please see the Appendix for additional baseline comparisons, analyses, and discussion of our limitations.

\vspace{-1mm}
\subsection{Evaluation Protocol} \label{sec:eval_protocol}
\vspace{-1mm}
We evaluate our method along two axes: adherence of the output image to each input visual prompt and the diversity of variations in the scene layout. 

\vspace{-2mm}
\myparagraph{Adherence to input.}
Compositional generation involves creating images with diverse scene layouts and poses, making it challenging to quantify how faithfully the output adheres to the input visual prompts. 
First, since composed images combine multiple prompts, measuring similarity between individual prompts and \emph{entire} output image is inappropriate, as it does not accurately reflect each prompt's contribution. 
Second, variations in pose and spatial arrangement, which we \emph{desire} in our output, can lower similarity scores even when object identities are preserved.

To address this, we propose a new \emph{compositional identity metric} that employs a feature extractor $F$. We first apply an open-set object detection algorithm to the generated images to identify candidate objects~\cite{ravi2024sam2, yuan2021florence}. We then extract features with $F$ for both the input and detected objects, capturing high-level semantic features robust to pose and layout changes. Using the Hungarian Algorithm with pairwise feature similarity as the cost function, we find an optimal matching between the input and detected objects. The final matching cost serves as our compositional identity metric.
Following previous works that measured identity preservation for the personalization task, we use both DINOv2~\cite{oquab2023dinov2} and CLIP~\cite{radford2021learning} as our feature extractors and denote the corresponding scores as $\text{DINO}_\text{comp}$ and $\text{CLIP}_\text{comp}$, respectively.
Notably, this metric naturally accounts for cases where prompted objects are missing or duplicated in the output.

\vspace{-6mm}
\myparagraph{Scene layout variations.}
To measure the diversity of layouts generated by each method, we produce five different output compositions from the same input visual prompts using different random seeds. Following previous works~\cite{NIPS2017_6650, zhao2021comodgan, Liu_DivCo}, we then compute the average LPIPS ~\cite{zhang2018perceptual} distance between each pair of these output images. A higher average LPIPS value indicates that the method generates more diverse images in response to varying random seeds.

\begin{table}[ht!]
    \centering
    \resizebox{1.0\linewidth}{!}{
    \begin{tabular}{l ccc}
        \toprule  \\
        \multirow{2}{*}{\makecell{\textbf{Method}}} &
        \textbf{Diversity} & \multicolumn{2}{c}{\textbf{Identity Preservation}}
        \\
        \cmidrule(lr){3-4}
        & $\text{LPIPS}_\text{avg}$ ($\uparrow$) 
        & $\text{DINO}_\text{comp}$ ($\uparrow$)
        & $\text{CLIP}_\text{comp}$ ($\uparrow$)
        \\
        \cmidrule(lr){1-4}

        IP-Adapter & 
        0.669 & 0.201 & 0.481\\
        
        IP-Adapter Plus & 
        0.578 & 0.255 & 0.560\\
        
        BLIP-Diffusion &
        \textbf{0.734} & 0.209 & 0.511 \\
        
        \hdashline
        
        KOSMOS-G & 
        0.687 & 0.294 & 0.596\\
        
        $\lambda$-ECLIPSE &
        0.671 & 0.241 & \underline{0.669}\\
        
        \hdashline
        Break-A-Scene & 0.587 & \underline{0.363} & 0.655\\
        \hdashline
        \textbf{\methodName (ours)} &  \underline{0.688} & \textbf{0.518} & \textbf{0.676}\\

        \bottomrule 
    \end{tabular}
    }
    \caption{\textbf{Quantitative comparisons.} We compare our method with prior image prompting, multi-modal generation, and optimization-based approaches. Output diversity is measured by $\text{LPIPS}_\text{avg}$ and identity preservation is measure through $\text{DINO}_\text{comp}$ and $\text{CLIP}_\text{comp}$. 
    The best result is marked in \textbf{bold}, and the second best is \underline{underlined}.
    }
    \label{tab:cmp_baselines_combined}
    \vspace{-2mm}
\end{table}

\vspace{-5mm}
\myparagraph{Evaluation datasets.} 
We adapt DreamBooth dataset~\cite{ruiz2022dreambooth}, originally containing single-object images, for our composition task. We randomly sample individual objects and combine them with a random background image, generated by applying an inpainting the images. 
We generate 300 inputs, each comprising 3, 4, or 5 visual prompts.

\subsection{Comparison to Existing Methods} \label{sec:cmp_baselines}
\vspace{-2mm}
Table~\ref{tab:cmp_baselines_combined} and Figure~\ref{fig:comparisons} compares \methodName with three families of prior approaches: image prompt methods~\cite{ye2023ipadapter, li2023blipdiffusion}, multimodal generative methods~\cite{pan2023kosmos, patel2024lambdaeclipse}, and an optimization based method~\cite{avrahami2023bas}. 

Since image prompt methods do not natively support multiple input images, we adapted them following community recommendations~\cite{diffusers}. For IP-Adapter~\cite{ye2023ipadapter}, we incorporated multiple images by summing the outputs of the decoupled cross-attention layers. For BLIP-Diffusion~\cite{li2023blipdiffusion}, we handled multiple images by averaging the tokens across input visual prompts.

IP-Adapter uses a coarse representation to encode inputs, which, as discussed in Section~\ref{sec:arch}, results in difficulties adhering to the input visual prompts. This is quantitatively observed in its low identity preservation score in Table~\ref{tab:cmp_baselines_combined} and visually seen in Figure~\ref{fig:comparisons}, where it fails to accurately generate the inputs.
In contrast, IP-Adapter Plus employs a fine-grained image encoder but struggles to generate diverse outputs. This limitation is reflected in the lack of diversity in the second column of Figure~\ref{fig:comparisons}, and the lower diversity score $\text{LPIPS}_\text{avg}$  in Table~\ref{tab:cmp_baselines_combined}. 
BLIP-Diffusion achieves a high diversity score but a low identity preservation score due to attribute bleeding, as illustrated in Figure~\ref{fig:comparisons}. Instead of generating two separate objects—a gray cat and a red chow chow dog—BLIP-Diffusion blends the concepts, producing a reddish gray cat.
All image prompt methods blend the identities of the objects.

KOSMOS-G~\cite{pan2023kosmos} and $\lambda$-ECLIPSE~\cite{patel2024lambdaeclipse} both struggle to generate multiple objects, resulting in low identity preservation scores. Their outputs also contain severe leaking of attributes. For instance, in the second example, KOSMOS-G generates a hybrid of a red vase and a rubber duck. 

Finally, we evaluate the optimization-based Break-A-Scene~\cite{avrahami2023bas}, which extracts multiple concepts from a single input image.
Since it requires all objects to appear in the same image, we adapted it for compositional generation by pasting object segments onto a background at random positions.
 Break-A-Scene struggles in for image composition because it relies on realistic interactions within the input image.
 As shown in Figure~\ref{fig:comparisons}, the generated outputs have minimal diversity in object poses and exhibit unnatural layouts—for instance, the cat in the first image and the duck in the second are floating mid-air with inconsistent shadows and lighting. Moreover, this method is computationally expensive due to its per-image optimization requirement.

Our method outperforms each prior method in terms of diversity of output generations as measured by $\text{LPIPS}_\text{avg}$ and the adherence to input visual prompts measured by $\text{DINO}_\text{comp}$ and $\text{CLIP}_\text{comp}$. Qualitative results in Figure~\ref{fig:comparisons} shows that all existing methods struggle to generate multiple objects in a realistic layout.

\subsection{Analysis} \label{sec:cmp_ablations}
Here we show the effectiveness of KV-Mixed cross attention and compositional guidance. Please refer to the Appendix for additional analysis results.  

\vspace{-2mm}
\myparagraph{KV-Mixed Cross-Attention.}
First, we analyze the importance of mixing keys and values discussed in \ref{sec:arch} by considering two settings: first uses \emph{only} coarse encoder for both keys and values, and second that uses \emph{only} fine-grained encoder. Figure~\ref{fig:plt_ablation} shows that using coarse encoder has poor identity preservation, indicated by low $\text{DINO}_\text{comp}$ scores, whereas using a fine-grained encoder has pood diversity, shown through lower $\text{LPIPS}_\text{avg}$ score.

\vspace{-3mm}
\myparagraph{Compositional Guidance.}
Our compositional guidance technique further improves both object identity preservation and the diversity of outputs.
Figure~\ref{fig:ablations} visually illustrates this enhancement, showing better adherence to the input visual prompts. For example, in the top image, there is minor attribute leakage where the stuffed bear is generated with the duck's beak. By applying compositional guidance, which restricts the attention maps, we reduce attribute leakage and enhance the identity preservation of the generated objects. Figure~\ref{fig:plt_ablation} validates the improvement quantitatively.

\section{Conclusion} \label{sec:conclusion}

We introduce a method for compositional image generation that integrates object-level visual prompts directly into the feed-forward process of image synthesis. Our approach is designed to balance adherence to these visual prompts with the generative model’s ability to produce a rich variety of compositions.
Compared to text-based prompting, visual prompt composition offers more precise control over the visual output—embodying the principle that ``an image is worth a thousand words''.

In general, text-to-image models tend to struggle with generating complex scenes containing multiple objects, making the task challenging not only due to the demands of identity preservation and compositional diversity. Nevertheless, as we demonstrate, the use of visual prompts facilitates the creation of such complex scenes.

\myparagraph{Acknowledgements.}
This research was performed while Gaurav Parmar was interning at Snap.
We thank Sheng-Yu Wang, Nupur Kumari, Maxwell Jones, and Kangle Deng for their fruitful discussions and valuable input which helped to improve this work.
We also thank Ruihan Gao and Ava Pun for their feedback regarding the manuscript. This work was supported by Snap Research, NSF IIS-2239076, and the Packard Fellowship.

{
    \small
    \bibliographystyle{ieeenat_fullname}
    \bibliography{main}
}
\clearpage

\appendix

\begin{figure*}[ht!]
    \centering 
    \includegraphics[width=\linewidth]{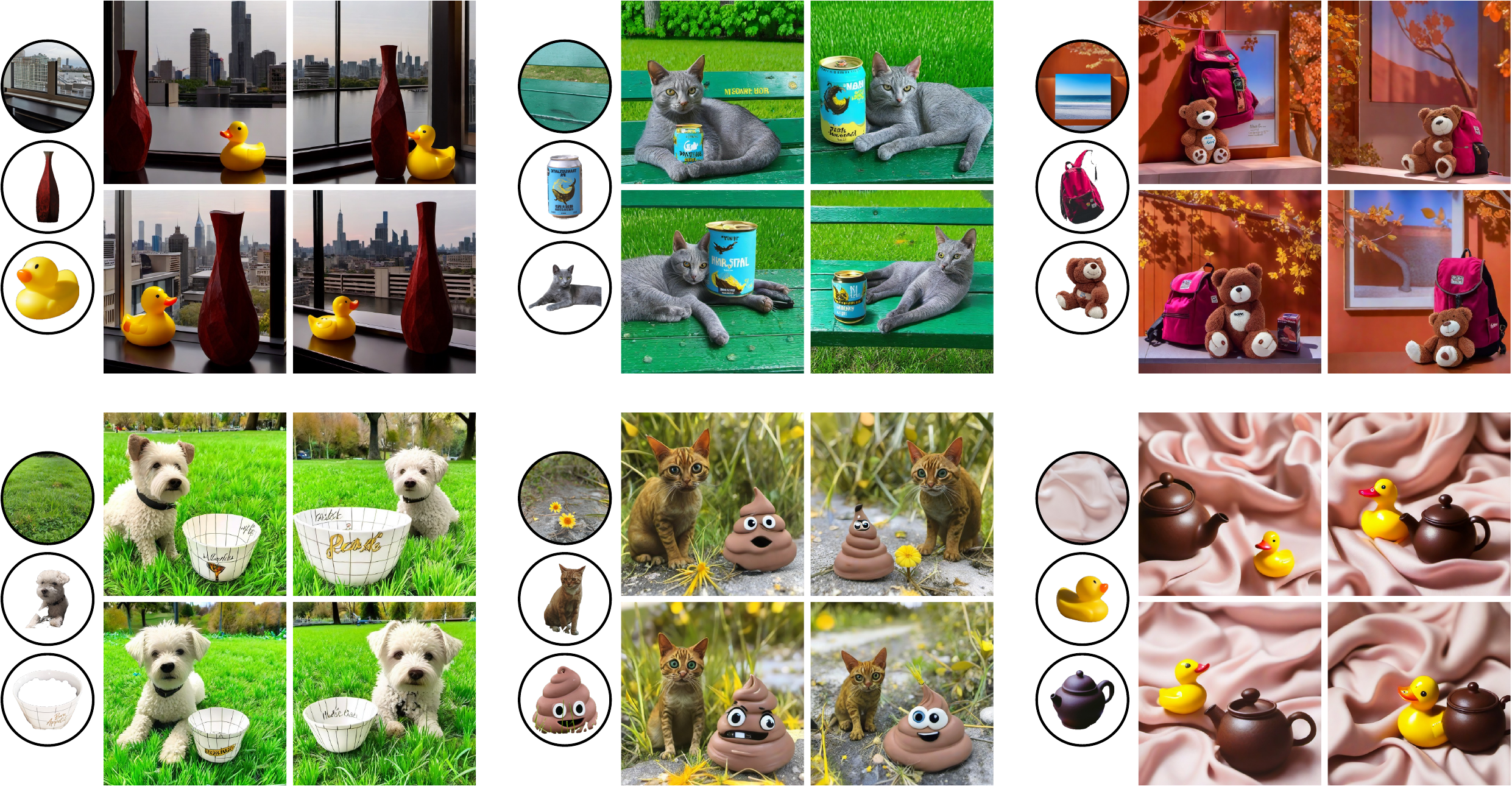}
    \caption{\textbf{Compositional generation results.} We show additional image composition results here. The input visual prompts are shown on the left and the generated compositional images are shown on the right.
    }
    \label{fig:sup_results_sdxl}
\end{figure*}

\begin{figure*}[ht!]
    \centering 
    \includegraphics[width=\linewidth]{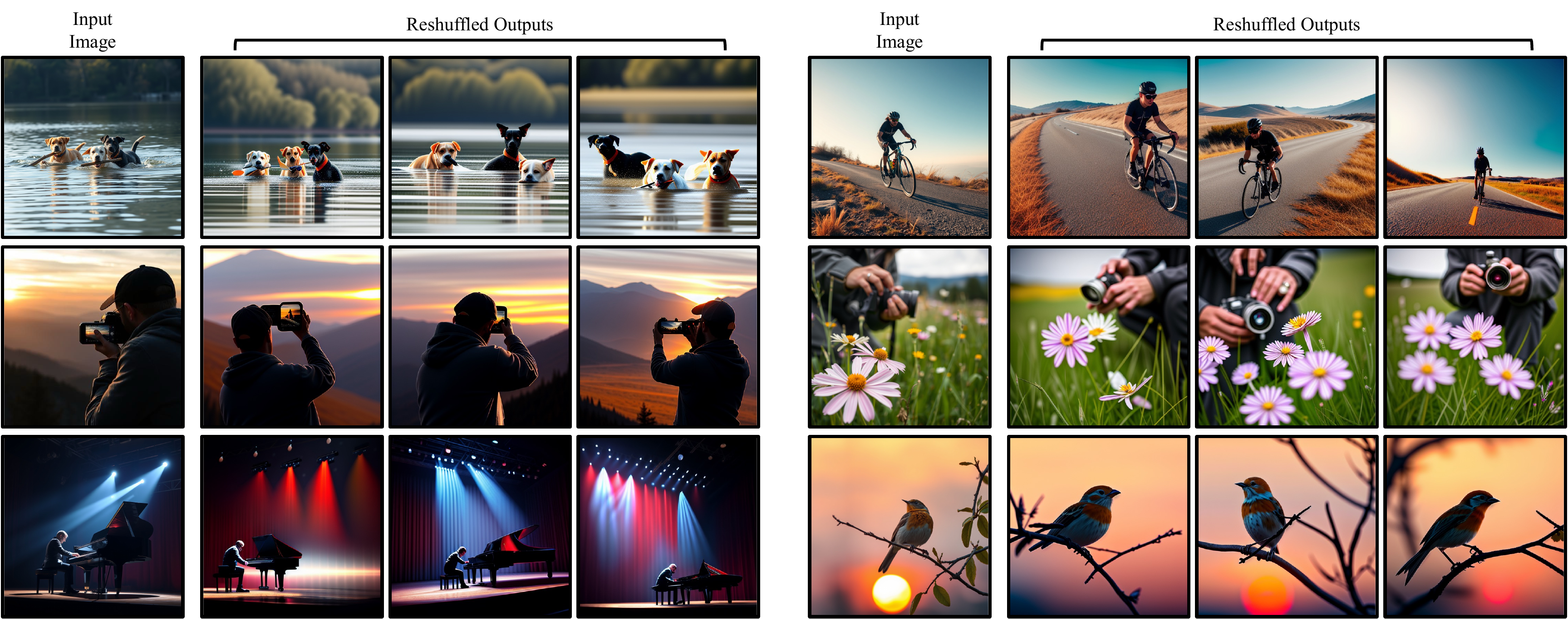}
    \vspace{-5mm}
    \caption{\textbf{Gallery of reshuffling results.} The input images is shown on the left and three reshuffled results are shown on the right. 
    }
    \label{fig:sup_results_reshuffling}
\end{figure*}

\begin{figure*}[ht!]
    \centering 
    \includegraphics[width=\linewidth]{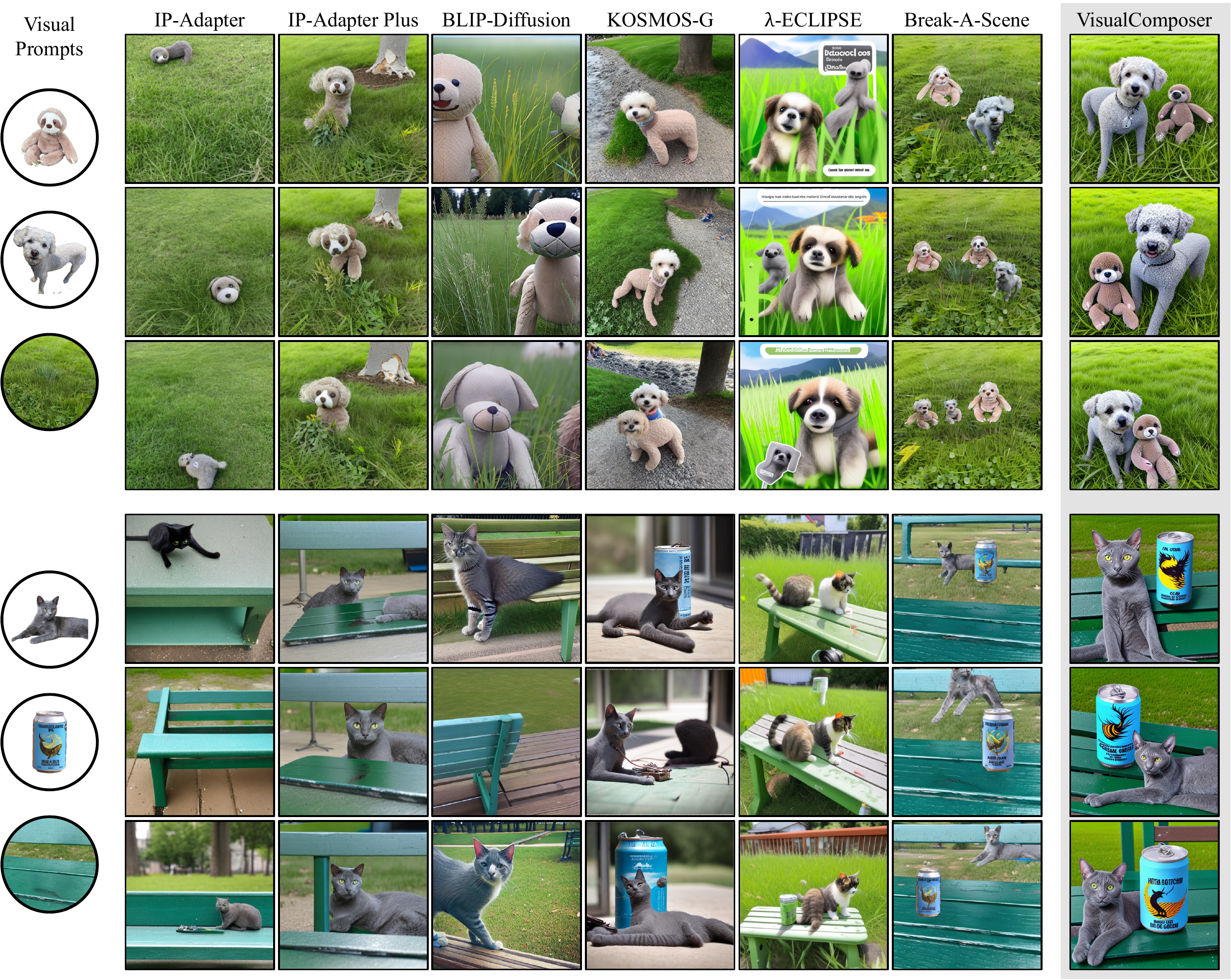}
    \vspace{-5mm}
    \caption{\textbf{Additional comparisons to prior methods.} We show a set of input visual prompts on the left. For each set, we show results generated by different methods. Our method outperforms each of the prior methods in terms adherence to the input visual prompt and diversity.
    }
    \label{fig:sup_comparisons}
\end{figure*}

\begin{figure*}[ht!]
    \centering 
    \includegraphics[width=\linewidth]{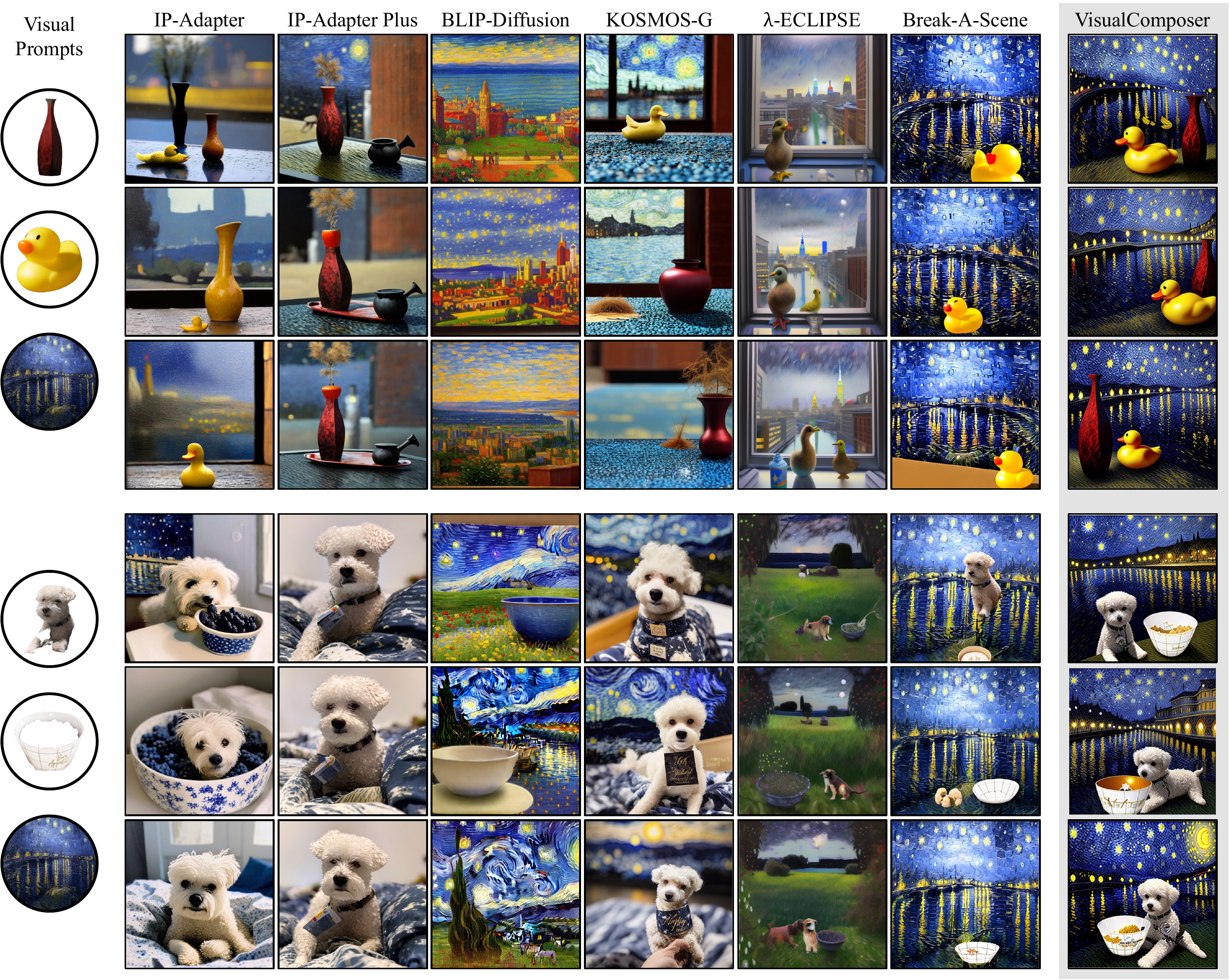}
    \vspace{-5mm}
    \caption{\textbf{Additional comparisons to prior methods with a painting background prompt.} We show a set of input visual prompts on the left. Notably, the background prompt for both examples is a painting.
    }
    \label{fig:sup_comparisons_paintings}
\end{figure*}

\begin{figure}[ht!]
    \centering 
    \includegraphics[width=\linewidth]{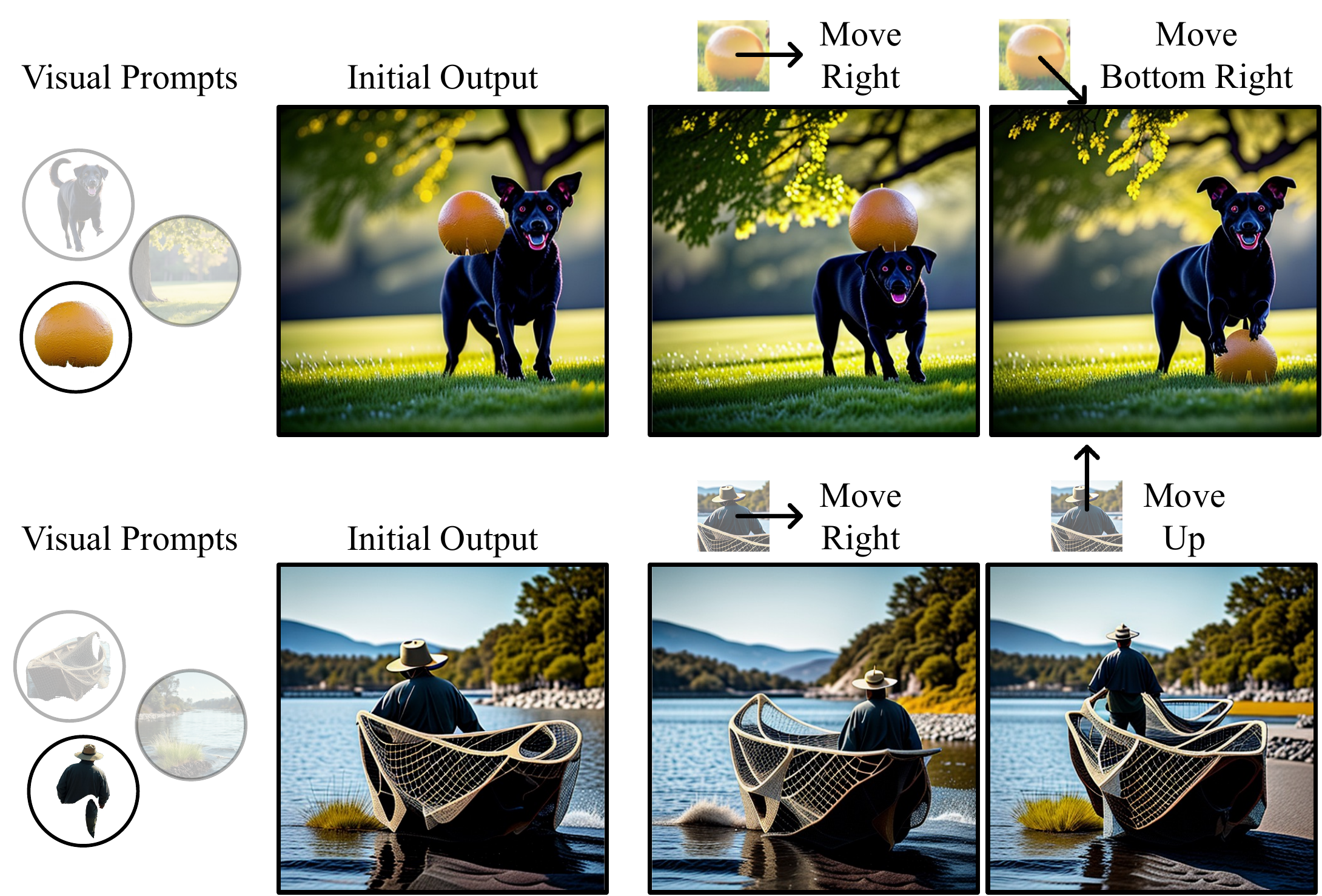}
    \caption{\textbf{Translation control.} Our object-level image prompts provide fine-grained control over each object. For example, we move the orange ball by manipulating its attention map, and the dog's pose changes in response to the ball's location.}
    \label{fig:sup_translation_control}
\end{figure}

\begin{figure}[ht!]
    \centering 
    \includegraphics[width=\linewidth]{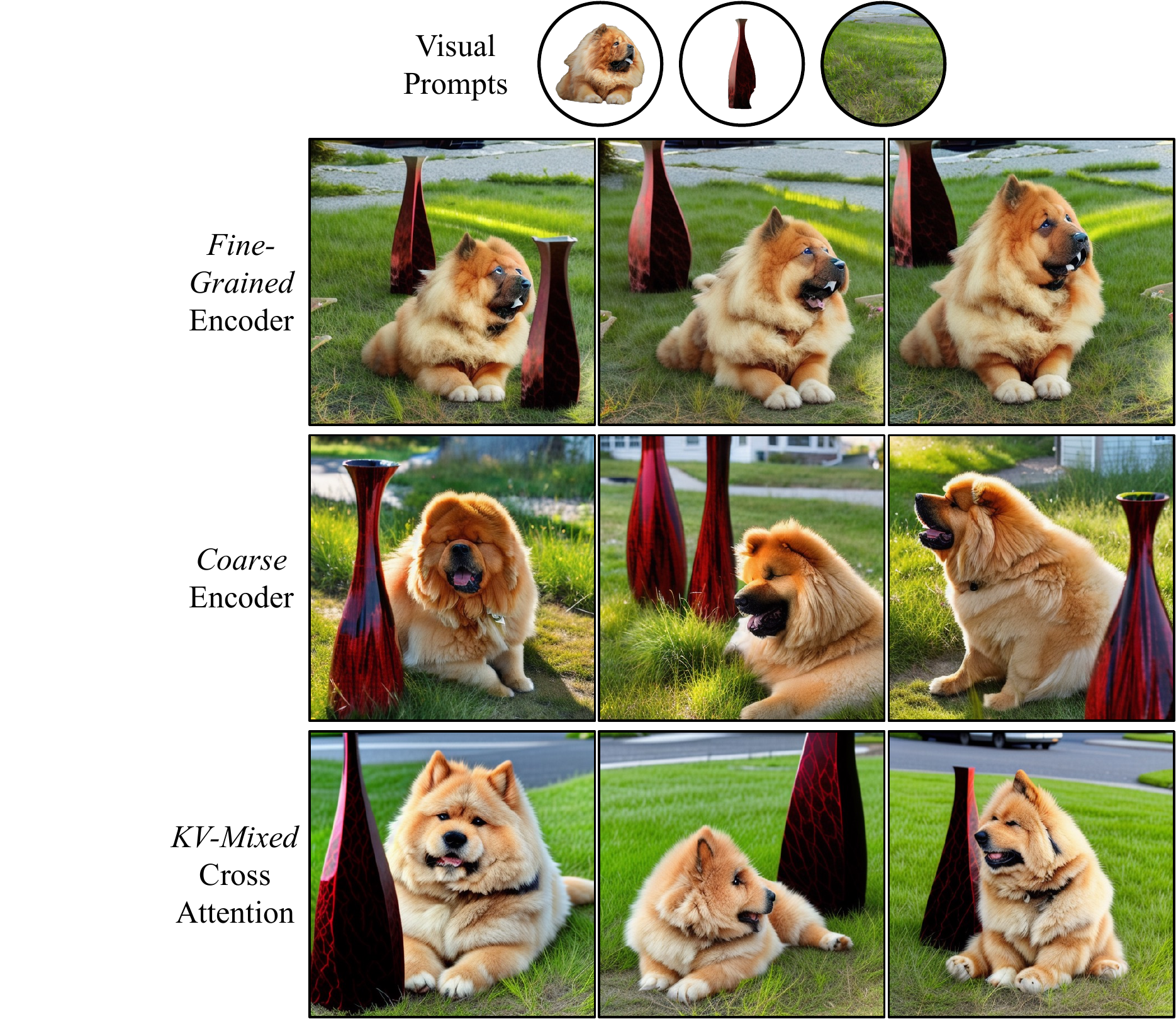}
    \caption{\textbf{Visual ablation of KV Mixture.} 
    }
    \label{fig:sup_kvmix_visual}
\end{figure}

\begin{table*}[ht!]
    \centering
    \begin{tabular}{l cc}
        \toprule 
        \multirow{2}{*}{\makecell{\textbf{Method}}} &
        \textbf{Ours} & \textbf{Baseline}
        \\
        & \textbf{Preferred}
        & \textbf{Preferred}
        \\
        
        \cmidrule(lr){1-3}
        \methodName (ours) vs IP-Adapter & \textbf{71.8\%} & 28.2\% \\
        \methodName (ours) vs IP-Adapter Plus & \textbf{59.9\%} & 40.1\% \\
        \methodName (ours) vs BLIP-Diffusion & \textbf{70.5\%} & 29.5\% \\
        \methodName (ours) vs KOSMOS-G & \textbf{62.1\%} & 37.9\% \\
        \methodName (ours) vs $\lambda$-ECLIPSE & \textbf{72.6\%} & 27.4\% \\

        \bottomrule 
    \end{tabular}
    \caption{\textbf{User Preference Study.} We evaluate adherence to the input visual prompts through a user study. Each comparison with a baseline comprises 13,500 questions asked to the users. Our results are preferred by users over those of each baseline. }
    \label{tab:supp_human_study}
\end{table*}

\begin{figure}[ht!]
    \centering 
    \includegraphics[width=\linewidth]{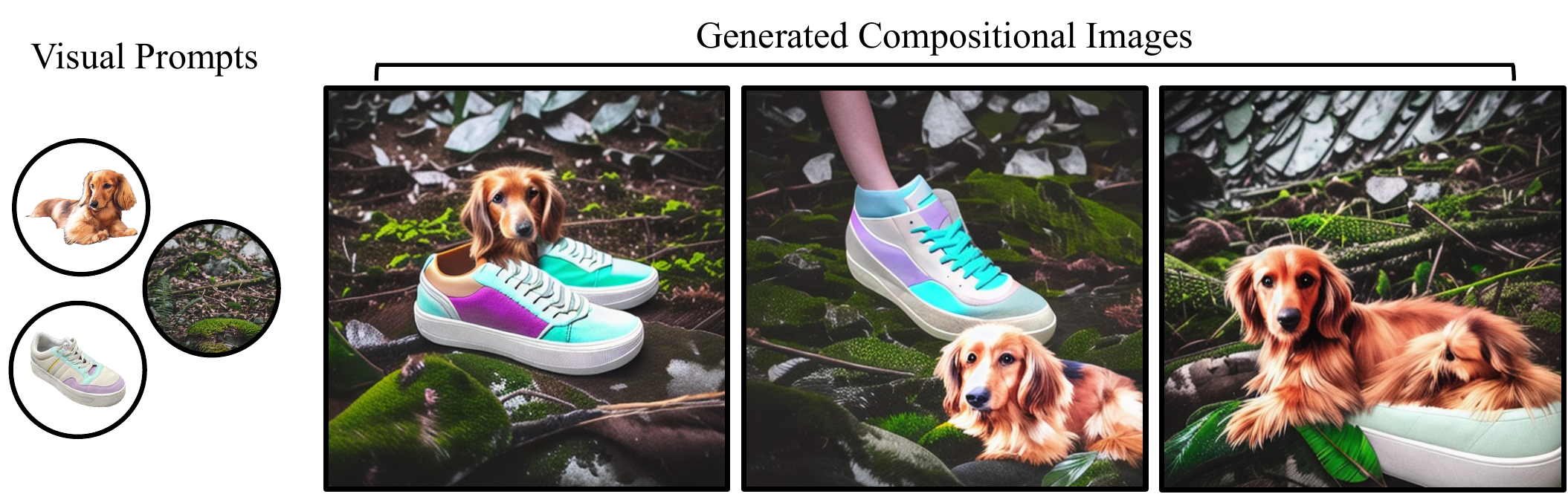}
    \caption{\textbf{Limitations.} We show the results of an example that illustrates the limitations of our method. Our method tends to perform worse for unusual combinations of input visual prompts.  
    }
    \label{fig:sup_limitations}
\end{figure}

Appendix~\ref{sec:sup_results} presents additional qualitative results obtained by our method. 
In Appendix~\ref{sec:sup_analysis}, we provide further analysis of different components of our method, followed by more comparisons with prior methods in  Appendix~\ref{sec:sup_comparisons}. 
Finally, Appendices~\ref{sec:sup_implementation} and \ref{sec:sup_limitations} include the implementation details and discuss the limitations of our method, respectively.

\section{Additional Results}
\label{sec:sup_results}

\myparagraph{Additional qualitative results.}
We show addition qualitative results in Figure~\ref{fig:sup_results_sdxl}. We use a classifier-free guidance scale of 5 for these results. 

\myparagraph{Reshuffling.}
Reshuffling is a special case of compositional generation where all input visual prompts are extracted from the same starting image. Figure~\ref{fig:sup_results_reshuffling} shows a large grid of reshuffling results generated by our method. 

\myparagraph{Object control.}
Our object-level cross-attention design enables precise control over individual objects in generated images. Figure~\ref{fig:sup_translation_control} illustrates this with two examples. In the first example, the input visual prompts are a dog, a grassy background, and an orange ball. The leftmost column shows the initial output image. By manipulating the cross-attention maps corresponding to the ball, we can move it above the dog's head (middle column) or near its lower right foot (right column). As the ball is repositioned, the scene adapts accordingly: the dog adjusts its pose by ducking its head when the ball is above it or standing on the ball when it's near its feet.

In the second example shown at the bottom, we change the position of a man standing in a boat. By moving him to the right, the reflection in the water adjusts accordingly. When moved upward, the man stands taller, revealing more of his legs. These examples demonstrate how our method allows for fine-grained control over object placement, with the scene naturally adapting to the changes.

\section{Analysis}
\label{sec:sup_analysis}
\myparagraph{KV-Mixed Cross-Attention.}
Figure 7 in the main paper quantitatively demonstrates the importance of KV-Mixed Cross-Attention layers, and Figure~\ref{fig:sup_kvmix_visual} visually illustrates their effects. 
The top row shows the results of using a fine-grained image encoder for both keys and values. This configuration causes the model to overfit to the poses of the input objects, producing outputs that closely mirror the input visual prompts. For example, the dog is always sitting in the same pose and looking to the right, identical to the input image.
In contrast, the middle row uses a coarse image encoder. Here, the generated images exhibit diverse poses and layouts, but the identities of the objects are not well preserved. For instance, the red vase looks different from the input, and the dog's fur does not match the original.
Finally, the bottom row illustrates the effects of our proposed KV-Mixed Cross-Attention. This approach enables us to generate diverse images while accurately retaining the identities of input visual prompts.

\section{Additional Comparisons}
\label{sec:sup_comparisons}
\myparagraph{Visual comparisons.}
Figure 5 in the main paper shows a visual comparison between our method and prior methods on two examples. In Figures~\ref{fig:sup_comparisons} and~\ref{fig:sup_comparisons_paintings}, we show additional visual comparisons.

\myparagraph{User preference study.}
We conduct a user preference study in addition to assessing adherence to the input visual prompts using automatic compositional identity metrics shown in the main paper Table 1 ($\text{DINO}\text{comp}$ and $\text{CLIP}\text{comp}$). Specifically, we perform pairwise preference comparisons in which users are shown three images: the input visual prompt and output images generated by two different methods. Users are then asked to choose which output images more accurately portray the input visual prompt.
Each comparison is performed by three different users, and a total of 13,500 comparisons are made for comparison with each of the baseline methods. The results in Table~\ref{tab:supp_human_study} show that our method is preferred over all prior encoder-based and multi-modal methods.

\section{Implementation Details}
\label{sec:sup_implementation}
\myparagraph{Dataset creation.}
As described in Section 3.3 of the main paper, our training dataset consists of images, their corresponding text prompts, a background image, and binary masks for individual objects and the background. The text prompts are generated automatically by recaptioning the images using LLaVa~\cite{liu2023improved}. To obtain precise binary segmentation masks, we first apply an open-set detection model~\cite{xiao2024florence} to identify bounding boxes within the images. We then use these bounding boxes to prompt SAM2~\cite{ravi2024sam2}, which provides accurate segmentation masks for each object. 
The background images are generated using the SD2.1 inpainting pipeline. 
We filter the dataset by discarding images that have a CLIP-Aesthetic score below 5.0, a minimum dimension (height or width) less than 512 pixels, or contain fewer than three or more than six objects.

\myparagraph{Training hyperparameters.}
We train all models using the Adam optimizer~\cite{kingma2014adam} with a learning rate of 0.0001 and a batch size of 32, for a total of 40,000 update steps on four NVIDIA A100 GPUs. To enable classifier-free guidance during inference, we randomly drop the text prompts and visual prompts during training: each is independently dropped 10\% of the time, and both are simultaneously dropped 5\% of the time.

\section{Limitations and Societal Impacts.}
\label{sec:sup_limitations}
We show the limitations of our model in Figure~\ref{fig:sup_limitations}. 
Our method has difficulty when users input combinations of visual prompts that are not commonly associated. For instance, in the figure, the input visual prompts include a dog, a single shoe, and a forest background. This unusual combination is challenging for our model, leading to failure cases,  such as hallucinating a leg wearing the shoe or generating an extra shoe.

Compositional image generation has the potential to democratize creative expression, allowing users to effortlessly synthesize complex scenes by assembling various visual elements. However, they also pose societal challenges, such as the risk of creating realistic but deceptive images that could spread misinformation or infringe on intellectual property rights. To counter these issues, it is important to explore detecting generated images~\cite{wang2020cnn} or attributing them corresponding source visual prompts~\cite{wang2024attributebyunlearning}.

\end{document}